\pdfoutput=1

\documentclass[11pt, UTF8]{article}
\usepackage{CJKutf8}
\usepackage{ulem}

\usepackage[final]{acl}

\usepackage{times}
\usepackage{latexsym}
\usepackage{microtype}
\usepackage{graphicx}
\usepackage{subfigure}
\usepackage{booktabs} %
\usepackage{amsmath}
\usepackage{amssymb}
\usepackage{mathtools}
\usepackage{amsthm}
\usepackage{hyperref}
\usepackage{multirow}

\usepackage[T1]{fontenc}

\usepackage[utf8]{inputenc}

\usepackage{microtype}

\usepackage{inconsolata}

\usepackage{framed}
\definecolor{formalshade}{rgb}{0.85, 0.95, 0.95}

\title{Sing it, Narrate it: Quality Musical Lyrics Translation}

\author{Zhuorui Ye$^{*}$, 
 Jinhan Li$^{*}$,  Rongwu Xu  \\
IIIS, Tsinghua University \\
\texttt{\{yezr21, lijinhan21, xrw22\}@mails.tsinghua.edu.cn}}

\begin{document}
\maketitle

\def\thefootnote{*}\footnotetext{Equal contribution.}\def\thefootnote{\arabic{footnote}}

\begin{abstract}
    Translating lyrics for musicals presents unique challenges due to the need to ensure high translation quality while adhering to singability requirements such as length and rhyme. Existing song translation approaches often prioritize these singability constraints at the expense of translation quality, which is crucial for musicals. 
This paper aims to enhance translation quality while maintaining key singability features. Our method consists of three main components. First, we create a dataset to train reward models for the automatic evaluation of translation quality. Second, to enhance both singability and translation quality, we implement a two-stage training process with filtering techniques. Finally, we introduce an inference-time optimization framework for translating entire songs.
Extensive experiments, including both automatic and human evaluations, demonstrate significant improvements over baseline methods and validate the effectiveness of each component in our approach. More results can be found in \href{https://lijinhan21.github.io/musical-translation/}{our website}. 

\end{abstract}

\section{Introduction}
    \begin{figure*}[h]
  \centering
  \includegraphics[width=1\linewidth]{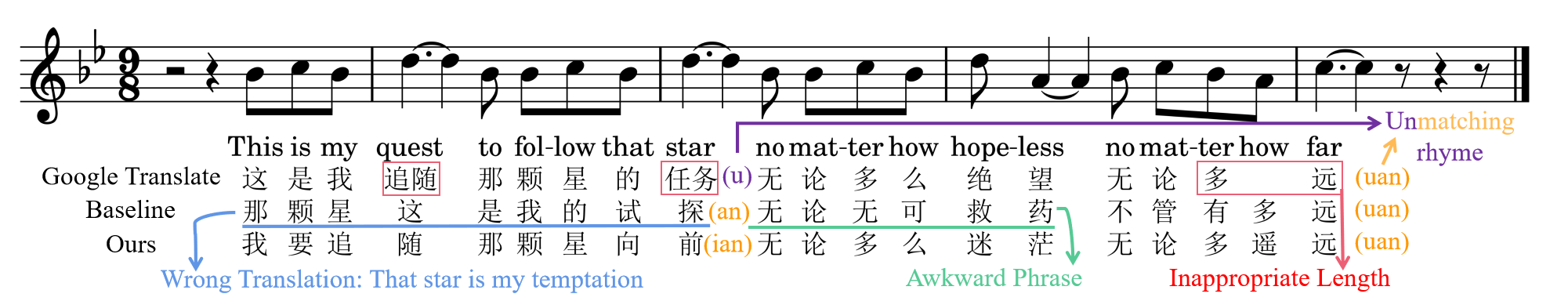}
  \vspace{-0.4cm}
  \caption{Aspects we considered include length, rhyme, and translation quality. The proper length of translated lyrics is the number of notes, and the end rhyme of each line (shown in parentheses) is better to have the same type (shown in the same color). Google translation fails to follow the length constraint and misaligns with music, as shown in red boxes, and its rhyme does not match. Both the baseline and our results meet length and rhyme constraints, but the baseline has inaccurate translations and inappropriate phrases, while our model generates higher-quality lyrics. }
  \label{fig:constraints}
\vspace{-0.2cm}
\end{figure*}

Have you ever heard of \href{https://www.bilibili.com/video/BV1YN411873B/}{Hamilton in Chinese}, or Mamma Mia in Swedish ~\cite{kerstrm2010TranslatingSL}?
Advancements in cultural globalization allow musicals to reach universal audiences, but language barriers still hinder full comprehension.
Translating musicals into performing country's language could enhance audience experience~\cite{sorby2014translating} and expand commercial outreach~\cite{andersson2008mamma}, as it allows audiences to enjoy theatrical elements without heavily relying on subtitles~\cite{engel2006words, sorby2014translating}.
However, musical translation is labor-intensive and time-consuming, requiring adjustments for musical framework, stage performance, and cultural references beyond mere verbatim translation~\cite{sorby2014translating, yuanhong2014fei}.
To alleviate this burden, we aim to automatically translate musical lyrics from English to Chinese.

Song translation, a closely related field, requires aligning the translated text with the music to ensure the translated lyrics can be sung~\cite{doi:10.1080/0907676X.2003.9961466, MusicalComedyTranslationFidelityandFormatintheScandinavianMyFairLady}. However, musical translation requires an even higher standard of translation quality because lyrics play a crucial role in the story-telling of a musical~\cite{kenrick2010musical, carpi2020multimodal, chan2017visible}. To preserve the original narration, the translations must accurately convey the meaning and nuance of the source lyrics. This high fidelity ensures that the translated musical maintains its artistic integrity and allows the story to unfold as intended for the target audience. Thus, musical translation demands a rigorous approach to translation quality, focusing on maintaining the narrative function to create a faithful rendition of the original work.

To the best of our knowledge, there is no existing work on automatic musical translation, and existing works on automatic song translation~\cite{guo-etal-2022-automatic, ou-etal-2023-songs, li-etal-2023-translate} mainly focus on the alignment of text and music, sacrificing translation quality and often produce unnatural and inaccurate translations unsuitable for musicals, as shown in Figure~\ref{fig:constraints}. 
To distinguish our work from existing art, we focus on improving translation quality, which would contribute to maintaining the narrative function, while reasonably following singability constraints. We define translation quality using the well-established criteria for literature translation: fluency, accuracy, and literacy~\cite{yan1898fu}. Additionally, we consider the singability constraints of length and rhyme following previous works~\cite{guo-etal-2022-automatic, ou-etal-2023-songs}. Figure~\ref{fig:constraints} shows our considered aspects, with examples demonstrating their significance. 

To depict translation quality, we collect some English-Chinese lyric pairs with human annotations. We scrape English musical lyrics and generate the corresponding Chinese lyrics by prompting large language models, label them according to our scoring rubrics, and train reward models to provide evaluations that correlate with human scores. 
For singability constraints, we observe that large language models struggle to adhere to them in a zero-shot manner. 
Thus, we perform two-stage translation model training to improve accuracy, balancing singability with translation quality using filtered high-quality data.
Finally, to produce coherent translations for entire passages, we propose an inference-time optimization framework that leverages the output diversity of large language models and a loss function designed to optimize paragraph-level overall quality. Extensive experiments demonstrate the effectiveness of our method's components, significantly outperforming the previous state-of-the-art approach.

To sum up, we make the following contributions:
(1) We propose the task of musical translation, which requires a higher level of translation quality than song translation;
(2) We create a dataset \texttt{MusicalTransEval} for scoring musical translation, which could serve as a valuable resource for future research; %
(3) We propose a two-stage translation model training approach that leverages reward models for data filtering and introduces a novel inference-time optimization framework, both aimed at improving translation quality while maintaining satisfactory singability performance.

\section{Related Work}
    \begin{figure*}[tp]
  \centering
  \includegraphics[width=0.9\linewidth]{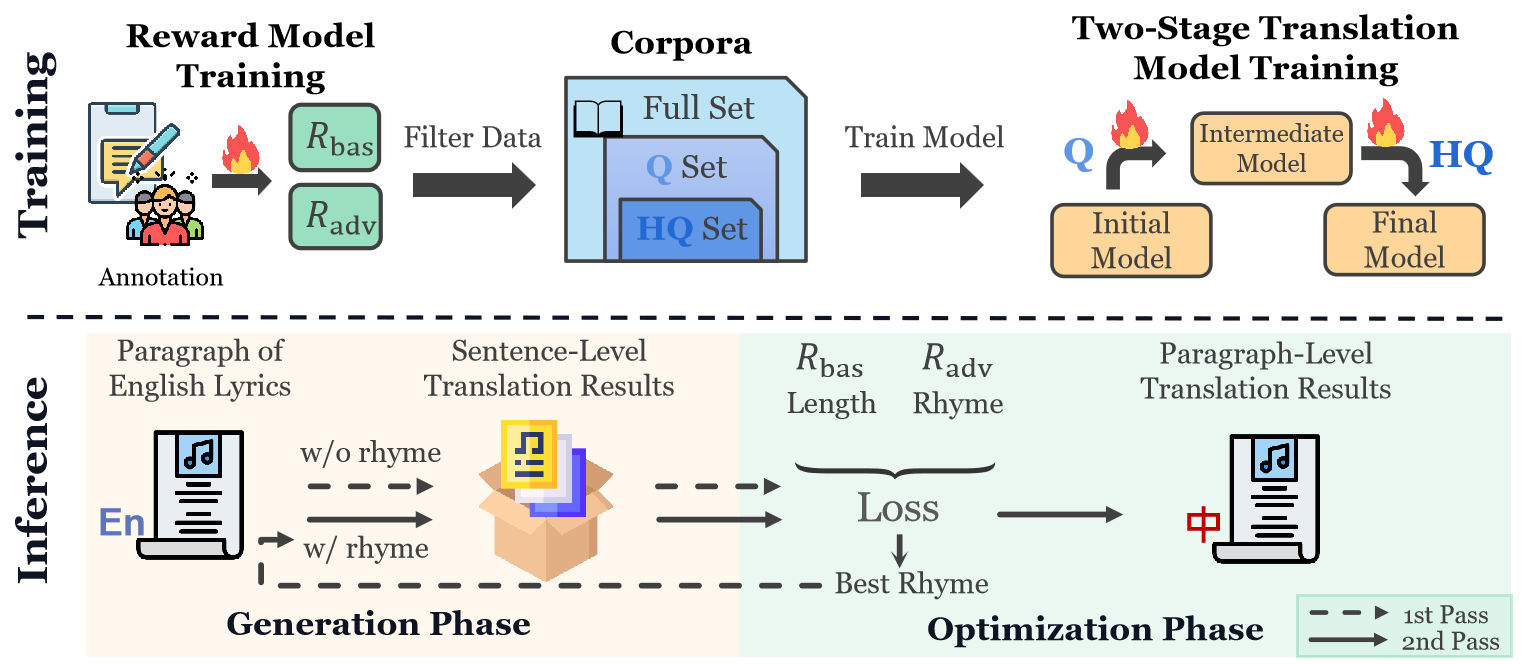}
  \caption{Overview of our pipeline. There are three key components in our method: reward model training (top left), translation model two-stage training (top right), and inference-time optimization framework (bottom). We use reward models to filter the whole corpora into a \textbf{Q}uality subset and a \textbf{H}igh-\textbf{Q}uality subset and train our generation model with the Q set and then with the HQ set. During inference, we generate plenty of sentence-level translations and derive paragraph-level translations by optimizing the loss function considering various aspects. We additionally give a 2nd pass with the same process but generate more sentence translations conditioned on the best rhyme.}
  \label{fig:pipeline}
\end{figure*}

\noindent\textbf{Translatology: Song and Musical Translation.}
In translatology, ``Pentathlon Principle'' ~\cite{doi:10.1080/0907676X.2003.9961466, 2005ThePA} is a well-known theory and guidance on general song translation~\cite{franzon2008choices, cheng2013singable, stopar2016mamma, si2017practical, opperman2018inter, sardina2021translation, pidhrushna2021functional, ou-etal-2023-songs}, which proposes five criteria to consider: singability, rhyme, rhythm, sense, and naturalness, where the first three relates to music-text alignment and the rest refer to translation quality. However, this principle is not developed specifically for songs on the musical stage and is not completely suitable for it~\cite{carpi2020multimodal}. 

The functional approach~\cite{MusicalComedyTranslationFidelityandFormatintheScandinavianMyFairLady} is more suitable for songs in musicals~\cite{carpi2020multimodal}, which emphasizes that the translated lyrics should replicate the function of the source text. In musicals, songs are ``story-telling'' elements~\cite{kenrick2010musical}, and the translated lyrics must carry out this role~\cite{desblache2018translation, kerstrm2010TranslatingSL, sorby2014translating,MusicalComedyTranslationFidelityandFormatintheScandinavianMyFairLady}. Thus a basic yet necessary constraint in musical translation is that lyrics must maintain the original narrative function, and thus should have high quality. 

\noindent\textbf{Automatic Song Translation.} 
To our best knowledge, there are only three previous works on automatic song translation~\cite{guo-etal-2022-automatic, ou-etal-2023-songs, li-etal-2023-translate}. \citet{guo-etal-2022-automatic} mainly addresses the problem of aligning words' tones with the melody in the beam search phase, and \citet{li-etal-2023-translate} focuses on aligning text to musical notes better. However, they both neglect the important rhyme constraint~\cite{10.1093/ml/II.3.211}. \citet{ou-etal-2023-songs} considers length, rhymes, and word boundaries, achieving decent results with prompting and the trick of reverse-order decoding. However, the translation quality is awkward and unsuitable for singing in musicals. To bridge this gap, we focus on generating high-quality translations under the two most important constraints for text-music alignment: length and rhyme.

\noindent\textbf{Large Language Model and Machine Translation.} 
Recent years have witnessed the huge success of large language models, including close-sourced GPT-4~\cite{openai2023gpt4}, \href{https://www.moonshot.cn/}{Kimichat}, and open-sourced Llama-2~\cite{touvron2023llama}. Recent works~\cite{yang2023bigtranslate,zhang2023bayling,zeng2024tim,chen2023improving,li2023eliciting,zhu2023extrapolating} sought to enhance the machine translation capability using open-sourced large models, yet the improvements are limited. One challenge is balancing performance improvements during fine-tuning without significantly compromising the pre-trained model's knowledge. As \citet{xu2024paradigm} pointed out, there is a diminished necessity for parallel data to fine-tune large language models, and it is recommended to first train with monolingual data if the language model does not have too much knowledge of the target language, and then fine-tune with a small amount of high-quality parallel data. Though our setting is slightly different, we similarly find it beneficial to fine-tune with high-quality parallel data.

\section{Problem Formulation}
    
We formulate the problem of musical translation as: Given a paragraph of English lyrics from a song, the task is to produce a Chinese translation that has high translation quality while adhering to singability constraints. By treating each paragraph independently, we can process an entire song.

To ensure \textbf{singability constraints}, we consider the following aspects.
(1) \textit{Length:} The number of syllables in the English lyrics and the number of characters in the Chinese lyrics should match the number of musical notes to ensure proper alignment. Since we lack direct access to sheet music but can easily obtain the English lyrics, we use the number of syllables in the English lyrics as the reference length for alignment.
(2) \textit{Rhyme:} The translated sentences within each paragraph should maintain the same end rhyme as much as possible, particularly aligning with the end rhyme of the last sentence in each paragraph.

To evaluate \textbf{translation quality}, we focus on the following three aspects~\cite{yan1898fu}.
(1) \textit{Fluency}: The naturalness and readability of the translated lyrics in Chinese.
(2) \textit{Accuracy}: How well the translation conveys the same meaning as the original English lyrics.
(3) \textit{Literary quality}: The aesthetic appeal and literary merit of the translated lyrics. 
We further categorize fluency and accuracy as basic translation quality, while considering literary quality as advanced translation quality, to differentiate between mandatory and supplementary aspects. To enable machines to evaluate these aspects of translation quality, we train reward models using human annotations as learning data.

\section{Method}

Our method consists of three key components: reward models trained to evaluate the quality of the translated language (Section \ref{sec32}), a translation model trained using a two-stage pipeline (Section \ref{sec33}), and an inference-time optimization framework that composes sentence-level results into paragraph-level output (Section \ref{sec34}). Figure \ref{fig:pipeline} illustrates how these components work together.

    \subsection{Reward Model Training}
        \label{sec32}
        
To train reward models for evaluating translations, we collect a dataset called \texttt{MusicalTransEval}, where each entry includes an original English line, a translated Chinese line, a paragraph as context, and three scores ranging from 1 to 4 that measure fluency, accuracy, and literacy of the translation respectively. The detailed scoring rubrics are shown in Appendix~\ref{appendixA}, which are developed in collaboration with an expert in musical translation. The English lines were extracted from musicals of diverse genres, ranging from fantasy, modern society, youth and family, history, and literature adaptation. The corresponding Chinese translations were generated by Kimichat using few-shot prompts. After 50 hours of annotation, we compiled a dataset with 3938 high-quality entries. 
For both basic and advanced translation quality, we train reward models using the dataset and refer to their evaluations as $R_{\mathrm{bas}}$ and $R_{\mathrm{adv}}$, respectively.

To obtain a more balanced training dataset for $R_{\mathrm{bas}}$ and $R_{\mathrm{adv}}$, we first apply mappings to handle categories that rarely appear. For $R_{\mathrm{bas}}$, we map the score pairs of fluency and accuracy to a single integer score ranging from 1 to 4, resulting in 471, 322, 971, and 2174 entries, respectively. For $R_{\mathrm{adv}}$, we map the scores for literacy to 2 or 3, obtaining 3104 and 834 data samples, respectively.

By utilizing data upsampling and downsampling techniques to further balance the training data, we obtained $R_{\mathrm{bas}}$ and $R_{\mathrm{adv}}$ with strong correlations with human judgments on a hidden balanced test set, which includes unseen musicals from the training period. The Pearson correlation~\cite{1895RSPS...58..240P} of human scores with $R_{\mathrm{bas}}$ and $R_{\mathrm{adv}}$ are 0.649 and 0.532, signifying strong and moderate correlation. Besides, the precision and recall of the score 3 class $R_{\mathrm{adv}}$ are 0.95 and 0.49. The strong correlation of $R_{\mathrm{bas}}$ and high precision of $R_{\mathrm{adv}}$ make them quite reliable and valuable in our pipeline. More details of \texttt{MusicalTransEval} can be found in Appendix \ref{appendixA} and more training details are in Appendix \ref{appendixB}.

    \subsection{Two-Stage Translation Model Training}
        \label{sec33}

Large-scale training is essential to ensure the translation model generates results that accurately adhere to length and rhyme constraints, as discussed in Section \ref{sec:analysis}. However, the same section also demonstrates that increasing the amount of training data does not always yield improvements in translation quality. This observation raises a pertinent question: how can we achieve high translation quality while maintaining satisfactory accuracy in terms of length and rhyme?

Due to the difficulty of collecting a large-scale musical dataset, we use the dataset provided by \citet{ou-etal-2023-songs}, consisting of approximately 2.8M English-Chinese song lyrics sentence translations. To bridge the gap between normal and musical songs and improve dataset quality, we use our reward models to filter a high-quality subset of 1.75M and a higher-quality subset of 700K entries.

In the first training stage, we train the LLM with the large-scale high-quality dataset to primarily learn to follow length and rhyme constraints. In the second stage, we further refine translation quality by fine-tuning with the higher-quality dataset. In both training stages, we use the same prompt with length and rhyme constraints, ensuring that the constraints-following ability learned in the first stage is maintained in the second stage. Additional descriptions of the training dataset can be found in Appendix \ref{appendixA} and more translation model training details are in Appendix \ref{appendixB}.

    \subsection{Inference-Time Optimization Framework}
        
\begin{table*}[t]
\centering
\fontsize{9.5}{10}\selectfont
\begin{tabular}{lccccccc}
\toprule
\textbf{Method (Training Config.)} & \textbf{Rhyme} & \textbf{LA} & \textbf{RS} & \textbf{$R_{\mathrm{bas}}$} & \textbf{$R_{\mathrm{adv}}$} & \textbf{BLEU} & \textbf{COMET} \\
\midrule
\citet{ou-etal-2023-songs} & yes & \textbf{0.977} & \textbf{0.96} & 2.845 & 2.053 & 18.01 & 71.94 \\
\multirow{2}{*}{Ours \textsc{Ver.1} (1.75M)} & yes & 0.941 & 0.722 & 2.789 & 2.046 & 18.22 & 71.93 \\
 & no & 0.854 & - & 2.92 & 2.053 & 17.15 & 71.61 \\
\multirow{2}{*}{Ours \textsc{Ver.2} (1.75M Q)} & yes & 0.914 & 0.687 & 2.971 & 2.056 & 18.32 & 72.87 \\
 & no & 0.819 & - & 3.063 & 2.059 & 17.68 & 72.49 \\
\multirow{2}{*}{Ours \textsc{Ver.3} (1.75M Q + 700K HQ)} & yes & 0.923 & 0.703 & \textbf{3.168} & \textbf{2.063} & 18.80 & \textbf{74.14} \\
 & no & \underline{0.874} & - & \underline{3.248} & \underline{2.068} & 17.76 & \underline{73.78} \\
\bottomrule
\end{tabular}
\caption{Sentence-level results of the three versions of our method. In \textsc{Ver.1}, we train the model with a 1.75M subset. In \textsc{Ver.2}, we use a 1.75M \textbf{Q}uality subset. In \textsc{Ver.3}, we use a 700K \textbf{H}igh-\textbf{Q}uality subset to fine-tune \textsc{Ver.2} model. Rhyme in the heading row means whether we use the rhyme constraint during inference, and the best results of the two cases are in \textbf{bold} (use) and \underline{underline} (without use), respectively.}
\label{table1}
\end{table*}

\begin{table*}[t]
\centering
\begin{tabular}{lcccccc}
\toprule
\textbf{Method (Training Config.)} & \textbf{LA} & \textbf{RS} & $R_{\mathrm{bas}}$ & $R_{\mathrm{adv}}$ & \textbf{BLEU} & \textbf{COMET} \\
\midrule
\citet{ou-etal-2023-songs} & 0.985 & \textbf{0.95} & 2.788 & 2.034 & 11.67 & 67.95 \\
Ours \textsc{Ver.1} (1.75M) & 0.988 & 0.806 & 3.608 & 2.243 & 10.39 & 69.42 \\
Ours \textsc{Ver.2} (1.75M Q) & 0.991 & 0.789 & 3.652 & 2.234 & 10.4 & 69.73 \\
Ours \textsc{Ver.3} (1.75M Q + 700K HQ) & \textbf{0.992} & 0.81 & \textbf{3.715} & \textbf{2.245} & 10.61 & \textbf{70.57} \\
\bottomrule
\end{tabular}
\caption{The final whole-song translation results of three versions of our method. Compared with Table \ref{table1}, our method includes the inference-time optimization framework here and can fully demonstrate our strength. }
\label{table2}
\end{table*}

        \label{sec34}
        Due to the inaccuracy of generating the whole paragraph at once, we let the translation model handle each sentence independently and then combine them using a novel optimization framework during inference. In particular, we design a proper paragraph-level loss function and optimize the overall loss by jointly considering all sentences. 

In our setting, we consider length accuracy, rhyme score, and both basic and advanced translation quality. At the paragraph level, our overall loss $\mathcal L(\cdot)$ is defined for sentence-level translations $y_1,\dots,y_n$ by incorporating all those aspects. Specifically, we define: \begin{align*}
    &\mathcal L(y_1,\dots,y_n)=\sum_i\left(\lambda_1 [\operatorname{Rhy}(y_i)\ne \operatorname{Rhy}(y_n)]\right.\\
    &\left.+\lambda_2 D(\mathrm{gt}_i,|y_i|)-\lambda_3 R_{\mathrm{adv}}(y_i)-\lambda_4 R_{\mathrm{bas}}(y_i)\right),
\end{align*}
where we define \begin{align*}
D(y,x) = 
\begin{cases} 
  \beta(x-y) & \text{if } y \le x, \\
  y-x & \text{if } y > x.
\end{cases}
\end{align*} 
to measure to which extent the translation length differs from the desired length, with an additional penalty $\beta$ for translations that exceed the desired length, as this poses a greater challenge for singing. The two reward models $R_{\mathrm{bas}}$ and $R_{\mathrm{adv}}$ are introduced earlier. $\operatorname{Rhy}(\cdot)$ specifies the rhyme type of the last character in a sentence, following the rhyme grouping rules from \citet{xue2002fan}, a Chinese music translation book. 
Additional details of the loss function are in Appendix \ref{appendixB}.

Our goal is then to find a paragraph translation that minimizes the optimization objective. We select an appropriate temperature for the generation function and generate a diverse set of candidate translations for each sentence to ensure a broad coverage of high-probability outputs in the generation space. This results in a vast number of possible combinations for $y_1,\dots,y_n$.
However, due to the structure of the optimization formula, we can solve it efficiently by first enumerating $\operatorname{Rhy}(y_n)$ for the last sentence, and then optimizing each sentence independently. It is worth mentioning that the flexibility of our proposed framework enables fine-grained control over additional singability constraints, which could be explored in future works.

After identifying the sentences $y_1,\dots,y_n$ that minimize the loss function, we set the corresponding end rhyme as the desired end rhyme. To ensure most sentences in a paragraph match the desired rhyme, we have another stage to generate additional samples for each sentence with rhyme conditioning. The second pass is more focused and sample-efficient, as the desired rhyme is fixed.

\section{Experiments}
    
In our experiments, we investigate the following research questions: 

\textbf{RQ 1} How well does our method perform in generating high-quality musical lyrics translations, as measured by automatic evaluation metrics? 

\textbf{RQ 2} How well do the generation results of our method align with human preference? 

\textbf{RQ 3} How does each component contribute to our performance improvements?

    \subsection{Experiment Configurations}
        
\begin{table*}[t]
\centering
\setlength{\tabcolsep}{5pt} %
\begin{tabular}{lcccccc}
\toprule
\multirow{2}{*}{\textbf{Method}} & \multicolumn{4}{c}{\textbf{Sentence-level}} & \multicolumn{2}{c}{\textbf{Paragraph-level}} \\
 & \textbf{Fluency} & \textbf{Accuracy} & \textbf{Literacy} & \textbf{Alignment} & \textbf{Quality} & \textbf{Alignment} \\
\midrule
\citet{ou-etal-2023-songs} & 2.88 & 2.53 & 2.37 & 2.48 & 2.08 & 2.92 \\
Ours \textsc{Ver.1} & 3.09 & 2.6 & 2.45 & 2.69 & 2.31 & 2.75 \\
Ours \textsc{Ver.2} & 3.25 & 2.64 & 2.54 & 2.6 & 2.27 & \textbf{2.98} \\
Ours \textsc{Ver.3} & \textbf{3.29} & \textbf{2.89} & \textbf{2.67} & \textbf{2.7} & \textbf{2.58} & 2.96 \\
\bottomrule
\end{tabular}%
\caption{Human evaluation of final whole-song translation results. Our three versions correspond to those shown in Table \ref{table1}, trained on different subsets: without filtering, with filtering, and with an additional second filtering.}
\label{tab:human-eval}
\end{table*}

\noindent\textbf{Datasets.} 
To evaluate musical translation performance, we additionally collect a dataset of English lyrics and quality Chinese translations from \href{https://music.163.com/}{Cloud Music}. This dataset includes 409 paragraphs and 1,742 lines from 56 popular songs of diverse musicals. We use this test set to evaluate both sentence-level translation and whole-song translation results.
More details can be found in Appendix \ref{appx:test_data}.

\noindent\textbf{Models.}
For both the generation model and the reward model, we choose \href{https://github.com/ymcui/Chinese-LLaMA-Alpaca-2}{\texttt{Chinese-Alpaca-2-13B}} ~\cite{Chinese-LLaMA-Alpaca} as our base model since it is pre-trained with a large amount of Chinese corpora and has satisfying instruction-following ability.

\noindent\textbf{Baselines.}
To the best of our knowledge, there are only three previous works on song translation, GagaST~\cite{{guo-etal-2022-automatic}}, Controllable Lyric Translation~\cite{ou-etal-2023-songs}, and LTAG~\cite{li-etal-2023-translate}. 
Due to data acquisition difficulties of GagaST and LTAG, we have \citet{ou-etal-2023-songs} as our baseline.
We train the baseline model directly using its \href{https://github.com/Sonata165/ControllableLyricTranslation}{released code}.

\noindent\textbf{Metrics.}
For automatic evaluation, we consider \textit{length accuracy (LA)}, defined as the percentage of translated sentences whose length equals the desired length (we set it as the length of reference translation for sentence-level testing, and as the number of syllables of the English lyrics for paragraph-level testing), \textit{rhyme score (RS)}, which is defined as the average percentage of sentences within each paragraph that exhibit identical end rhymes, \textit{basic and advanced translation quality $R_{\mathrm{bas}}$ and $R_{\mathrm{adv}}$} as defined in Section \ref{sec32}, statistic machine translation metric \textit{BLEU}~\cite{10.3115/1073083.1073135}, and model-based machine translation metric \textit{COMET} (we use the \texttt{Unbabel/wmt22-comet-da} variant).~\cite{rei-etal-2022-comet}.  
One caveat of BLEU is that it entirely depends on lexical form match and is sensitive to paraphrasing. On the other hand, COMET is robust and aligns much better with humans. COMET ranked 2nd in its alignment with humans among 20 metrics studied in~\citet{freitag-etal-2022-results}, while BLEU only ranked 19th. Thus we mainly use COMET as the machine translation metric and report BLEU scores only for completeness.

    \subsection{Automatic Evaluations}

The sentence-level performance of our generation models trained with several different recipes is reported in Table \ref{table1}. In this experiment, we consider sentences in a paragraph as independent ones and set the desired length and rhyme according to our reference translation. We find that our dataset filtering strategy can largely improve translation quality by increasing all of $R_{\mathrm{bas}}$, $R_{\mathrm{adv}}$, and COMET. Also, after deleting the rhyme constraint in the prompt during inference time, generation results are still satisfactory even with slight improvements of $R_{\mathrm{bas}}$ and $R_{\mathrm{adv}}$, though COMET slightly drops, partially due to the loss of length accuracy and therefore more misalignment with reference translation.

In this work, we focus more on the paragraph-level translation results shown in Table \ref{table2}, which again indicates that our training strategy is effective and both our two training stages can boost performance. Comparing our final results with the baseline's results, it is evident that we have achieved significant improvements across the majority of metrics. The only metric that ours is not as good as the baseline is the rhyme score since \citet{ou-etal-2023-songs} uses its reversed decoding technique to benefit rhyme following at the cost of language quality, but our rhyme score is already high enough for most applications, especially considering that even English lyrics in a paragraph does not guarantee the same rhyme. We thus answer \textbf{RQ 1} affirmatively: our method can indeed achieve much better translation quality while maintaining satisfactory singability performance.

    \subsection{Human Evaluations}

\begin{table*}[]
\centering
\resizebox{\textwidth}{!}{%
\begin{tabular}{ccc}
\toprule
\textbf{Original lyrics} & 
\textbf{\citet{ou-etal-2023-songs}} &
\textbf{Ours \textsc{Ver.3}} \\
\midrule
You are sixteen going on seventeen & \begin{CJK}{UTF8}{gbsn}你是\underline{十六个十七岁}\end{CJK} & \begin{CJK}{UTF8}{gbsn}你十六岁快要十七\end{CJK}  \\
Fellows will fall in line & \begin{CJK}{UTF8}{gbsn}伙伴们会结队\end{CJK} & \begin{CJK}{UTF8}{gbsn}兄弟们排成排\end{CJK}  \\
Eager young lads and rogues and cads& \begin{CJK}{UTF8}{gbsn}\underline{渴望}年少顽童和\underline{部队}\end{CJK} & \begin{CJK}{UTF8}{gbsn}\uwave{年少轻狂}的无赖痞子\end{CJK}  \\
Will offer you food and wine & \begin{CJK}{UTF8}{gbsn}献给你餐酒一杯\end{CJK} & \begin{CJK}{UTF8}{gbsn}会为你提供美食\end{CJK}  \\
\hline

Sing once again with me, &
\begin{CJK}{UTF8}{gbsn}再和我一起唱\end{CJK} & 
\begin{CJK}{UTF8}{gbsn}和我再一起唱 \end{CJK}\\

our strange duet, &
\begin{CJK}{UTF8}{gbsn}\underline{陌生}的重唱\end{CJK} & 
\begin{CJK}{UTF8}{gbsn}怪异对唱 \end{CJK}\\

my power over you, &
\begin{CJK}{UTF8}{gbsn}我对你的\underline{力量}\end{CJK} & 
\begin{CJK}{UTF8}{gbsn}我对你的控制 \end{CJK}\\

grows stronger yet &
\begin{CJK}{UTF8}{gbsn}更加\underline{茁壮}\end{CJK} & 
\begin{CJK}{UTF8}{gbsn}越来越强 \end{CJK}\\

\hline

Just because you find that life's not fair, & 
\begin{CJK}{UTF8}{gbsn}只因你发现生活不公平\end{CJK} & 
\begin{CJK}{UTF8}{gbsn}只因为你发现生活不公\end{CJK}\\

it doesn't mean that you just have to grin and bear it! & 
\begin{CJK}{UTF8}{gbsn}不代表只需要笑着忍痛 \end{CJK}& 
\begin{CJK}{UTF8}{gbsn}不等于只能强颜而忍耐 \end{CJK}\\

If you always take it on the chin and wear it & 
\begin{CJK}{UTF8}{gbsn}如果总是\underline{把它戴在你的头顶} \end{CJK}& 
\begin{CJK}{UTF8}{gbsn}如果总是\uwave{硬着头皮}强忍下来 \end{CJK} \\

Nothing will change. & 
\begin{CJK}{UTF8}{gbsn}不会变更\end{CJK} & 
\begin{CJK}{UTF8}{gbsn}永不更改 \end{CJK}\\

\bottomrule
\end{tabular}%
}
\caption{Qualitative results for our model and the baseline. Translational errors and awkward phrases are \underline{underlined}. Excellent lyrics are \uwave{underwaved}.}
\label{tab:qualitative-baseline}
\end{table*}

\begin{table*}[!h]
\centering
\setlength{\tabcolsep}{3pt}
\resizebox{\textwidth}{!}{%
\begin{tabular}{cccc}
\toprule
\textbf{Original lyrics} & 
\textbf{Ours \textsc{Ver.1}} &
\textbf{Ours \textsc{Ver.2}} &
\textbf{Ours \textsc{Ver.3}} \\
\midrule

Suddenly I'm flying company chaters &
\begin{CJK}{UTF8}{gbsn}忽然间我\underline{飞去}公司包机了\end{CJK} &
\begin{CJK}{UTF8}{gbsn}突然间我飞着公司的包机\end{CJK} & 
\begin{CJK}{UTF8}{gbsn}突然间我正坐着包机\underline{飞往}\end{CJK} \\

Suddenly everything's high &
\begin{CJK}{UTF8}{gbsn}突然什么都高涨\end{CJK} & 
\begin{CJK}{UTF8}{gbsn}突然什么都高涨\end{CJK} & 
\begin{CJK}{UTF8}{gbsn}突然一切都高涨
\end{CJK} \\

Suddenly there's nothing in between me and the sky & 
\begin{CJK}{UTF8}{gbsn}突然之间\underline{没有了我和天空相隔}\end{CJK} & 
\begin{CJK}{UTF8}{gbsn}突然之间\underline{隔着我和天空的天际}\end{CJK} & 
\begin{CJK}{UTF8}{gbsn}突然之间我和天空之间\uwave{无屏障}\end{CJK}\\
\bottomrule
\end{tabular}%
}
\caption{Qualitative results for our three versions. They are trained on different subsets: without filtering, with filtering, and with an additional second filtering. Translational errors and awkward phrases are \underline{underlined}. Excellent lyrics are \uwave{underwaved}.}
\label{tab:qualitative-training}
\end{table*}

We recruit 4 musical enthusiasts from our university to do the human evaluation. We randomly sample 30 sentences and 12 paragraphs from our test set, let baseline and different versions of our model generate 120 sentences and 48 paragraphs, and ask another musical enthusiast to sing all generated results out. Subsequently, we let the evaluators assign scores on fluency, accuracy, literacy, and music-text alignment for sentence results, and overall translation quality and music-text alignment for paragraph results. We provide detailed scoring rubrics with examples and require the participants to adhere to our rules.

The human evaluation results are shown in Table~\ref{tab:human-eval}. They are generally consistent with our automatic evaluations. The clear improvement of our \textsc{VER.1} over the baseline and the improvement of our \textsc{VER.3} over the previous two versions demonstrate the effectiveness of our inference-time optimization and training dataset filtering. We thus answer \textbf{RQ 2} affirmatively: our method can align well with human preference and achieve better human evaluation scores.

We also note that although our rhyme accuracy is not as high as \citet{ou-etal-2023-songs}, our singability scores in human evaluation are consistently higher than the baseline, indicating our rhyming accuracy is already good enough for human listeners. People might pay more attention to how we can hear the words clearly in the lyrics given music which could explain why we are seeing slightly improved results in text-music alignment. More details of human evaluation can be found in Appendix~\ref{appendixD}.

    \subsection{Qualitative Results}

\begin{table}[]
\centering
\resizebox{\columnwidth}{!}{%
\begin{tabular}{lccccccc}
\toprule
\textbf{Method} & \textbf{LA} & \textbf{RS} & $R_{\mathrm{bas}}$ & $R_{\mathrm{adv}}$ & \textbf{BLEU} & \textbf{COMET} \\
\midrule
GPT-4o 0-shot & 0.286 & 0.425 & 3.342 & 2.073 & 17.69 & 74.67 \\
GPT-4o 5-shot & 0.302 & 0.287 & 3.493 & 2.098 & 15.24 & \textbf{74.98} \\
Ours \textsc{Ver.3} & \textbf{0.992} & \textbf{0.81} & \textbf{3.715} & \textbf{2.245} & 10.61 & 70.57 \\
\bottomrule
\end{tabular}%
}
\vspace{-0.1cm}
\caption{Whole-song results derived by directly prompting GPT-4o to generate paragraphs with 0-shot or 5-shot prompts. }
\vspace{-0.3cm}
\label{tab:gpt-4o}
\end{table}

In this section, we show a few representative qualitative results, with more results in Appendix \ref{appendixC}. For all Chinese translations, the translation errors and awkward phrases are underlined, and the excellent lyrics are underwaved. 

Table~\ref{tab:qualitative-baseline} shows generation results of \citet{ou-etal-2023-songs}, and our model. In our selected examples, the baseline has nearly perfect length and rhyme, but its translation quality is bad, with about one-third of incorrect or awkward phrases. In comparison, the generation results of our model have perfect length accuracy and satisfactory rhyme score, and their translation results are fluent, correct, and sometimes impressive. Table~\ref{tab:qualitative-training} demonstrates the effectiveness of our training recipe. With further fine-tuning with high-quality data, the percentage of awkward phrases is reduced and more excellent translations emerge. 

    \subsection{Understanding the Contribution of Each Component}

\begin{table}[]
\centering
\resizebox{\columnwidth}{!}{%
\begin{tabular}{lccccccc}
\toprule
\textbf{Samples} & \textbf{LA} & \textbf{RS} & $R_{\mathrm{bas}}$ & $R_{\mathrm{adv}}$ & \textbf{BLEU} & \textbf{COMET} \\
\midrule
1 & 0.891 & 0.387 & 3.103 & 2.084 & 12.25 & 70.28 \\
80 & \textbf{0.998} & \textbf{0.839} & 3.71 & \textbf{2.282} & 10.8 & 70.46 \\
40+40 & 0.992 & 0.81 & \textbf{3.715} & 2.245 & 10.61 & \textbf{70.57} \\
\bottomrule
\end{tabular}%
}
\vspace{-0.1cm}
\caption{Comparison of no sampling, one-stage sampling, and our two-stage sampling strategy performance. 40+40 means the number of samples in two stages.}
\vspace{-0.3cm}
\label{tab:optimization-framework}
\end{table}

\begin{table}[]
\centering
\resizebox{\columnwidth}{!}{%
\begin{tabular}{lcccccc}
\toprule
\textbf{Reward} & \textbf{LA} & \textbf{RS} & $R_{\mathrm{bas}}$ & $R_{\mathrm{adv}}$ & \textbf{BLEU} & \textbf{COMET} \\
\midrule
no & \textbf{0.999} & \textbf{0.876} & 2.974 & 2.073 & 11.03 & 68.62 \\
yes & 0.992 & 0.81 & \textbf{3.715} & \textbf{2.245} & 10.61 & \textbf{70.57} \\
\bottomrule
\end{tabular}%
}
\vspace{-0.1cm}
\caption{The comparison of whether there are reward model terms in the inference loss function, signified by Reward in the heading row.}
\vspace{-0.3cm}
\label{tab:reward}
\end{table}

\begin{figure*}[]
    \vspace{-0.3cm}
	\centering
    \includegraphics[width=1\linewidth]{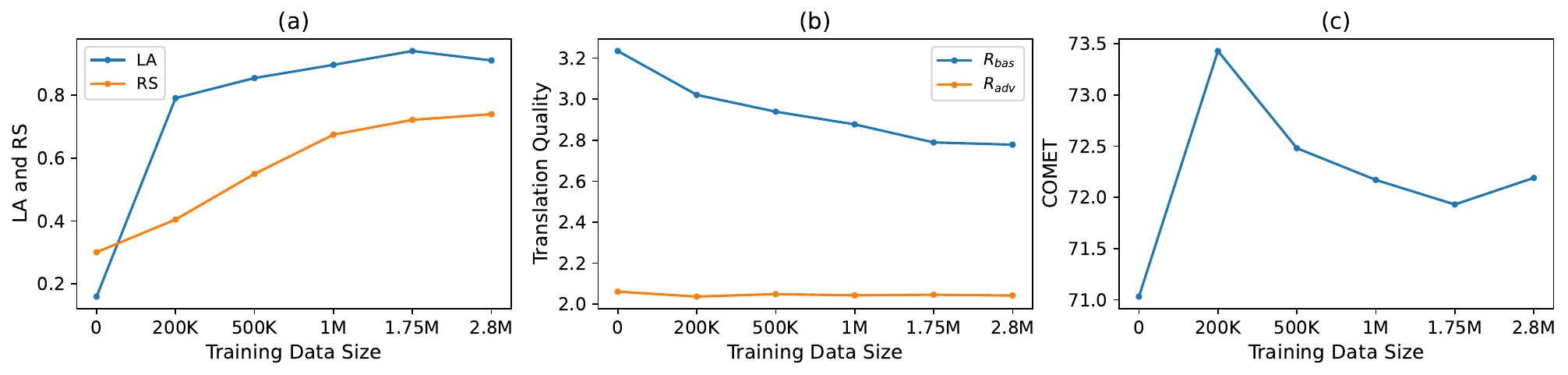}
    \vspace{-0.6cm}
	\caption{The changes of length accuracy, rhyme score, both basic and advanced translation quality, and COMET score if we change the training set scale. }
    \vspace{-0.2cm}
	\label{fig:sampling_number}
\end{figure*}

To answer \textbf{RQ 3}, we investigate the individual contribution of each component in our pipeline to the overall performance improvement.

\noindent\textbf{The necessity of fine-tuning translation models.}
We identified a key challenge of directly prompting one of the most advanced LLMs to translate lyrics. According to Table~\ref{tab:gpt-4o}, GPT-4o is unable to follow the length and rhyme constraints with only $0.302$ length accuracy and $0.425$ rhyme score, even with carefully designed few-shot prompting. These indicate the necessity of training a smaller model. The prompt for GPT-4o is in Table~\ref{tab:gpt-paragraph-prompt} in Appendix.

\noindent\textbf{Effectiveness of the optimization framework.} Table \ref{tab:optimization-framework} demonstrates the effectiveness of our optimization framework. If we forgo the optimization during inference and only rely on a single sampling step to obtain the final result, we observe significant drops across all metrics, particularly in the rhyme score. Interestingly, compared to a simple one-pass strategy with equal computational resources that only uses ensembling to fit a rhyme for a paragraph, incorporating a second stage does not give a better rhyme score with more rhyme-conditioned samples. We hypothesize it is because the trained model cannot perfectly guarantee rhyme following, thus some sentences could not fit the chosen rhyme.

\noindent\textbf{Impact of reward model terms in the inference loss.} We additionally demonstrate that incorporating reward model terms in the inference-time loss is critical to the overall performance improvement. Under our best-performing configurations, removing the reward model terms from the optimization process results in a decrease of more than 2 points in the COMET score, as shown in Table \ref{tab:reward}.
Compared to the one-sample setting in Table \ref{tab:optimization-framework}, the absence of reward model terms leads to a larger drop in the COMET score, as the model attempts to optimize the rhyme score at the expense of translation quality.
\noindent\textbf{Decomposing the sources of improvement.} Compared to the work of \citet{ou-etal-2023-songs}, while achieving comparable performance in terms of singability aspects, we analyze that the improvement in translation quality (approximated by the COMET score) can be primarily attributed to two factors. First, conducting dataset filtering using our trained reward models contributes to an improvement of approximately 1 points in the COMET score, as evidenced by Tables \ref{table1} and \ref{table2}. Additionally, the inclusion of reward model terms in the loss function of our inference-time optimization framework provides a further improvement of 2 points in the COMET score, as shown in Tables \ref{tab:reward}.

    \subsection{Additional Analyses}
        \label{sec:analysis}

\begin{table}[]
\centering
\resizebox{\columnwidth}{!}{%
\setlength{\tabcolsep}{5pt} %
\begin{tabular}{lccccccc}
\toprule
\textbf{Model} & \textbf{Trained} & \textbf{LA} & \textbf{RS} & $R_{\mathrm{bas}}$ & $R_{\mathrm{adv}}$ & \textbf{BLEU} & \textbf{COMET} \\
\midrule
Ours & no & 0.844 & 0.574 & 3.731 & 2.159 & 12.49 & 68.1 \\
Ours & yes & \textbf{0.99} & \textbf{0.873} & 3.76 & 2.248 & 12.32 & 69.43 \\
Kimichat & no & 0.944 & 0.669 & \textbf{3.777} & \textbf{2.271} & 15.98 & \textbf{72} \\
\midrule
Ours (new) & yes & 0.992 & 0.81 & \textbf{3.715} & \textbf{2.245} & 10.61 &                                                      \textbf{70.57} \\
Kimichat & no & 0.932 & 0.711 & \textbf{3.751} & \textbf{2.213} & 12.74 & \textbf{74.01} \\
GPT-4o & no & 0.984 & 0.929 & \textbf{3.75} & \textbf{2.192} & 12.97 & \textbf{73.67} \\
\bottomrule
\end{tabular}%
}
\vspace{-0.1cm}
\caption{The comparison of closed-sourced Kimichat and both our untrained and trained model variants. }
\vspace{-0.2cm}
\label{tab:kimi}
\end{table}

\begin{table}[]
\centering
\resizebox{\columnwidth}{!}{%
\begin{tabular}{lcccccc}
\toprule
\textbf{Samples} & \textbf{LA} & \textbf{RS} & $R_{\mathrm{bas}}$ & $R_{\mathrm{adv}}$ & \textbf{BLEU} & \textbf{COMET} \\
\midrule
10+10 & 0.979 & 0.652 & 3.66 & 2.184 & 11.62 & \textbf{71.29} \\
20+20 & 0.985 & 0.729 & 3.701 & 2.216 & 11.8 & 71.22 \\
40+40 & 0.992 & 0.81 & 3.715 & 2.245 & 10.61 & 70.57 \\
80+80 & \textbf{0.993} & \textbf{0.878} & \textbf{3.732} & \textbf{2.286} & 10.39 & 70.61 \\
\bottomrule
\end{tabular}%
}
\vspace{-0.2cm}
\caption{Comparison of different numbers of samples in our framework, all using two sampling stages.}
\vspace{-0.3cm}
\label{tab:samping-number}
\end{table}

\noindent\textbf{Impact of training data scale.} Figure \ref{fig:sampling_number} illustrates that increasing the scale of training data can help balance translation performance with length accuracy and rhyme score. Without training, the translation model struggles to adhere to length and rhyme constraints. As we increase the size of the training set, length and rhyme accuracy consistently improve, albeit at the cost of a slight drop in translation performance. This is expected, as our training helps the model follow the constraints but with imperfect translations, potentially diluting some of the pre-trained knowledge. To strike a balance, we use 1.75M data points to ensure high length and rhyme accuracy in the first training stage, and then employ high-quality filtered data to further refine translation quality in the second stage.

\noindent\textbf{Effect of sample count in our framework.} The number of samples used in our framework can be freely adjusted. As shown in Table \ref{tab:samping-number}, increasing the number of samples improves the rhyme score. In our pilot study, we find that using 40 samples for both the first and second stages can achieve a good balance between overall performance and computational efficiency. This setting takes about 1 minute for each paragraph, which is acceptable in terms of the real-world musical lyrics translation application.

\section{Conclusion}

In conclusion, our work successfully balances translation quality and singability in musical lyrics translation. To solve this task, we leverage trained reward models, a two-stage translation model training approach, and an inference-time optimization framework. Our approach ensures that translated lyrics meet the criteria of fluency, accuracy, and literary quality while adhering to the critical constraints of length and rhyme. The substantial improvements over the baseline, as evidenced by both automatic metrics and human evaluations, demonstrate the efficacy of our method in delivering high-quality translations that retain the essence of musical expression. This work paves the way for future advancements in the field, and advances the cross-cultural appreciation of musicals.

\section*{Limitations}
    Although the current version of our reward models can already achieve good results, there is room for further improvement by scaling the collected dataset and inviting more annotators to score sentence translations for less noise. We believe the results of the proposed method can be more impressive if we can access more resources to train better reward models.

Besides, we are translating at the sentence level due to the difficulty of tackling various constraints and composing sentences into a paragraph. Yet in some cases, neighboring sentence translations are not that compatible. Thus to further improve translation quality, we believe it is a promising direction to explore how to directly translate a paragraph.

Finally, in this work, we only consider two of the most critical singability aspects for simplicity. In future works, it is possible to consider more fine-grained singability constraints to make our compositions more professional. 

\section*{Ethics Statement}
    This work addresses the task of musical translation, considering both translation quality and singability constraints. Potential risks include inaccurate translation results, which may lead to misunderstandings if used directly in certain scenarios.

The lyric data used in this research are sourced from the public \href{https://music.163.com/}{Cloud Music platform} and are used solely for research purposes. The models are obtained from public GitHub repositories. The dataset provided by \citet{ou-etal-2023-songs} is also used in accordance with its original intended purpose. 

For human evaluations, we strictly adhere to the \href{https://www.aclweb.org/portal/content/acl-code-ethics}{ACL Code of Ethics}. Comprehensive details, including the recruitment process for evaluators and the instructions provided, are included in Appendix \ref{appendixD}. We collect evaluation scores without any personal information and ensure that the questionnaires do not contain offensive statements. Although our institute does not have an ethical review board or similar entity from which we can obtain approval, we have made every effort to follow the ethical guidelines set forth by ACL.

Regarding the use of AI assistants in our research, we primarily employed them for language polishing and refining the clarity of our writing. The main ideas, methodologies, and contributions presented in this paper are the result of our own work and intellectual efforts.

\section*{Acknowledgements}
    
We would like to thank Zhilin Yang, He Cheng, Yang Gao, Shengjie Wang, and Chonghua Liao for their insightful suggestions and help for this work. We thank all the musical enthusiasts from Tsinghua University who help us conduct human evaluations and provide us with valuable feedback.

\newpage
\normalem
\begin{CJK}{UTF8}{gkai}
\bibliography{custom}
\end{CJK}

\newpage
\appendix

\section{Dataset details}
    \label{appendixA}
    \subsection{\texttt{MusicalTransEval} Dataset for Reward Model}
\label{appx:reward_model}

For the \texttt{MusicalTransEval} dataset, we picked 11 musicals across various genres and spent 20 hours extracting all the lyrics from their songs, breaking them down into paragraphs. The distribution of the musical genres is shown in Figure~\ref{fig:annotation_distribution}(a). Next, we used the Kimichat API to get initial translations for these paragraphs, tweaking our pipeline a bit: we kept the optimization but focused only on length and rhyme scores, as we did not have reward models yet. 
We then labeled 3938 lines in three different aspects, which took us another 30 hours. We divided the labeled data into training and test sets. Time and budget constraints meant we could not label everything, but what we did manage to label already gave us pretty good results. 

\begin{figure*}[ht]
  \centering
  \includegraphics[width=1\linewidth]{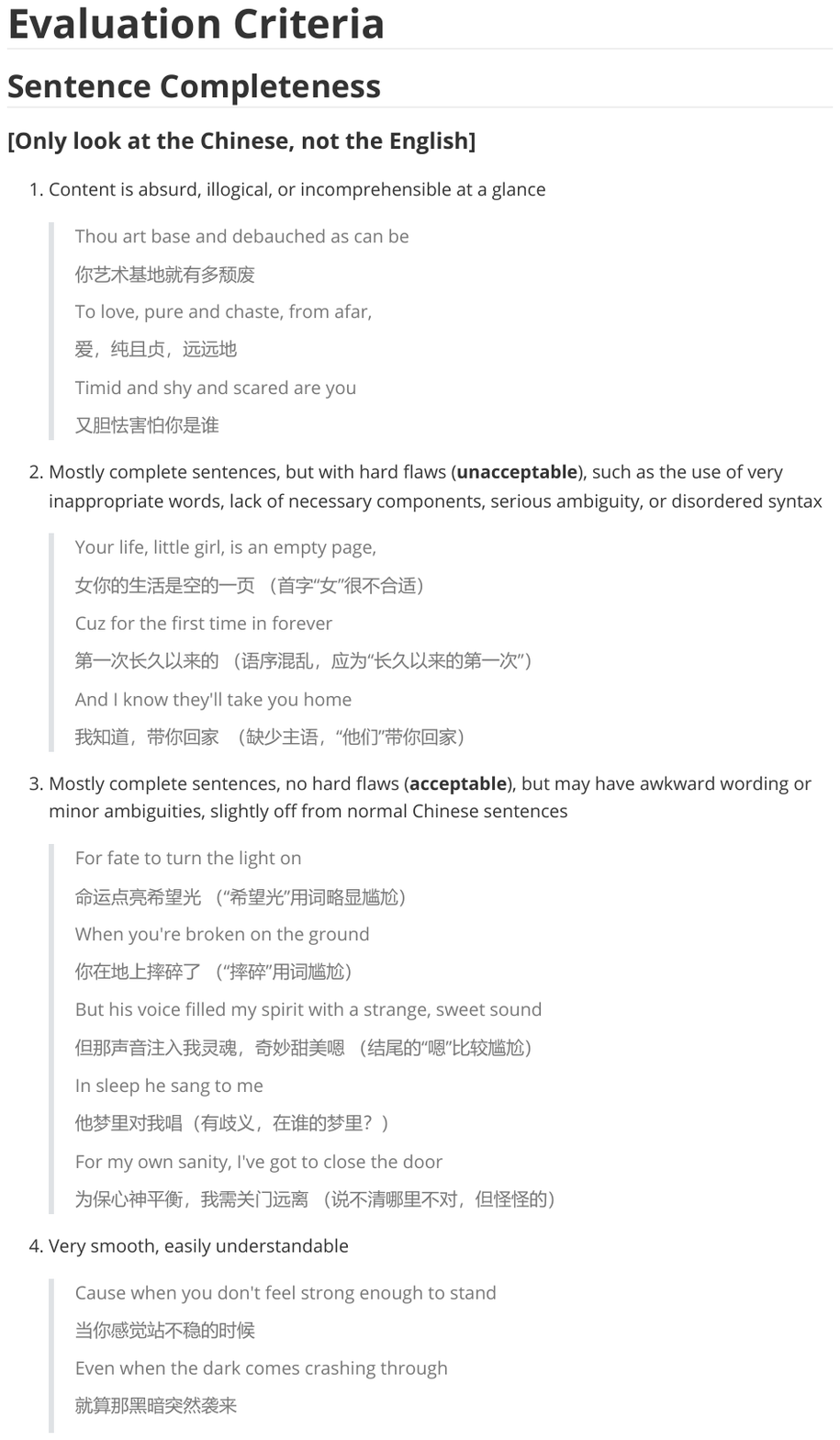}
  \vspace{-0.1cm}
  \caption{Metrics for human labeling, page 1/3.}
  \label{fig:reward1}
\vspace{-0.1cm}
\end{figure*}

\begin{figure*}[ht]
  \centering
  \includegraphics[width=1\linewidth]{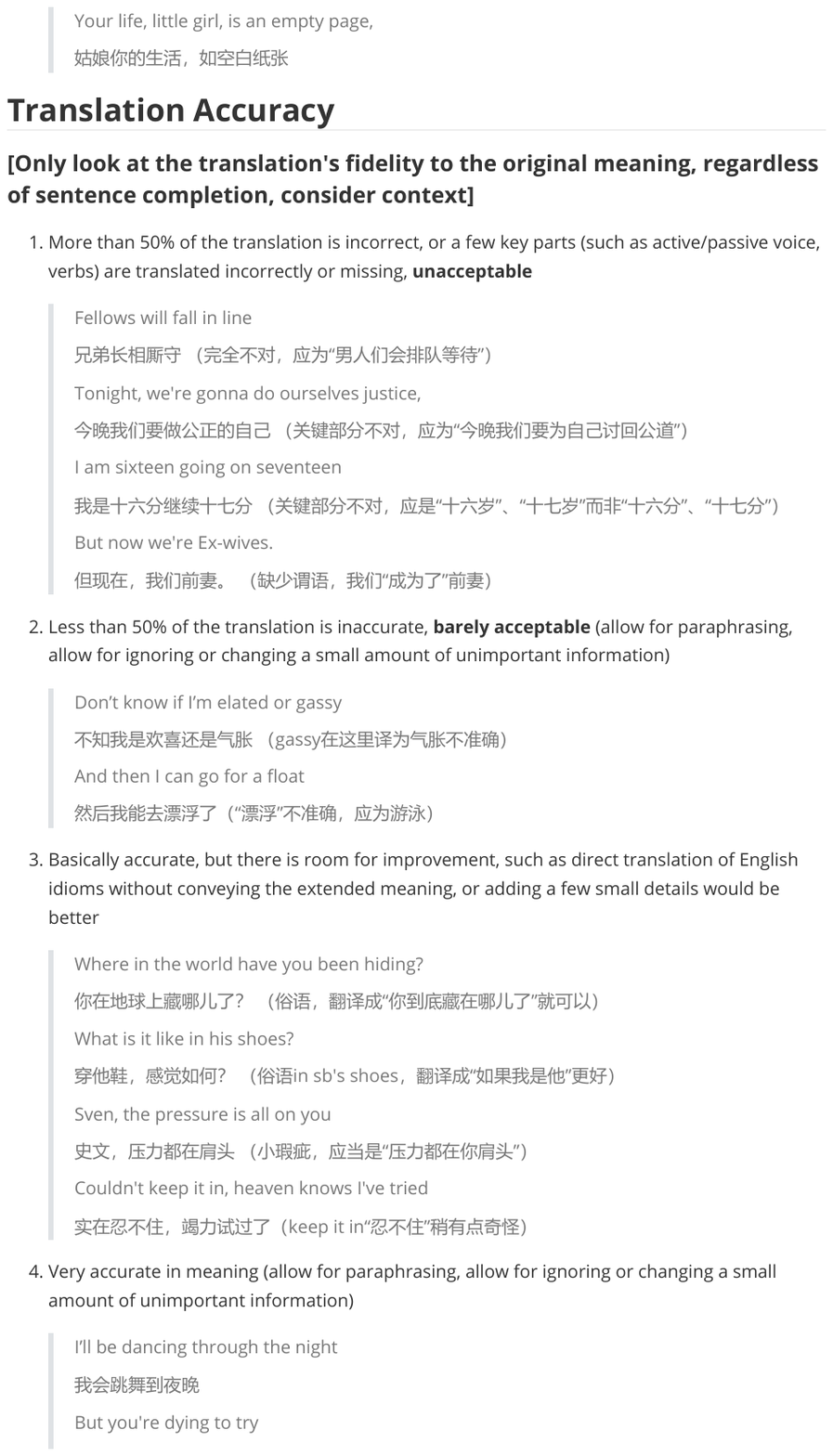}
  \vspace{-0.1cm}
  \caption{Metrics for human labeling, page 2/3.}
  \label{fig:reward2}
\vspace{-0.1cm}
\end{figure*}

\begin{figure*}[ht]
  \centering
  \includegraphics[width=1\linewidth]{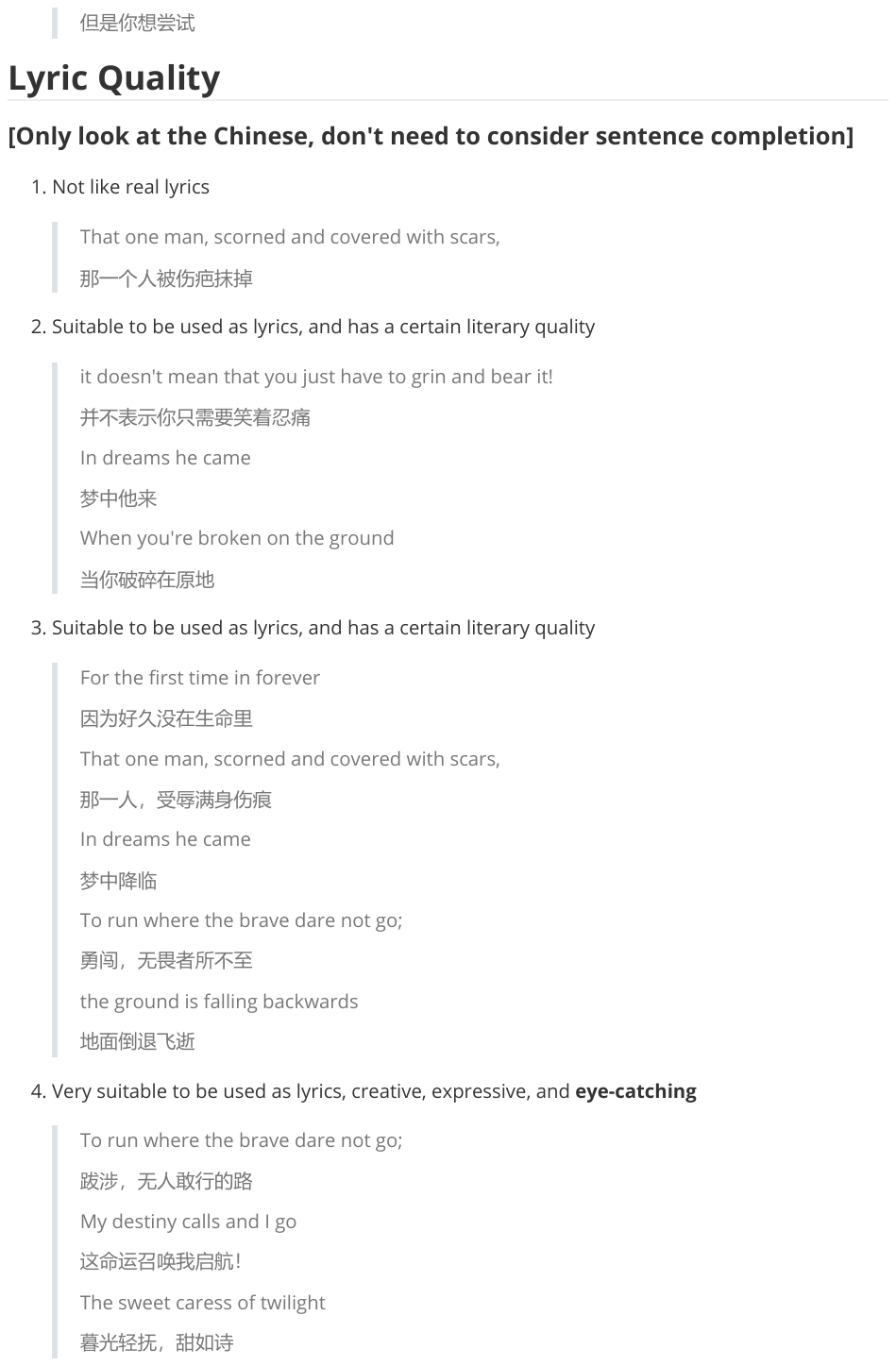}
  \vspace{-0.1cm}
  \caption{Metrics for human labeling, page 3/3.}
  \label{fig:reward3}
\vspace{-0.1cm}
\end{figure*}

Our labeling metrics is shown in Figure~\ref{fig:reward1},~\ref{fig:reward2},~\ref{fig:reward3}. 
We let human label in three aspects: fluency, translation accuracy, and literary. Each aspect has 4 levels of scores, and we give instructions and examples for each level to ensure consistency among human scores. We have endeavored to ensure a scientific and rational scoring process, collaborating with domain experts to establish sound criteria that have gone through a few amendments during the preliminary labeling stage. Also, we ensure annotators have a good background of musicals and are familiar with the rubrics, thus trying our best to reduce bias in annotations.

\subsection{Translation Model Training Dataset}
\label{appx:train_data}
As mentioned in Section~\ref{sec32}, due to the difficulty of collecting a large-scale musical dataset, we use the dataset provided by \citet{ou-etal-2023-songs}, which consists of approximately 2.8M song lyric sentence translations from English to Chinese for training. Although there is some gap between normal songs and musical songs, we bridge this gap and improve dataset quality by using our reward models to filter a high-quality subset of 1.75M and a higher-quality subset of 700K entries. The high-quality subset is obtained by selecting entries with a basic reward score $R_{\mathrm{bas}} \ge 3$, while the higher-quality subset is derived by choosing entries with $R_{\mathrm{bas}} = 4$. We observe that filtering the dataset using only the basic reward model already leads to improvements in the generated output. However, additionally employing the advanced reward model for filtering may result in overfitting, causing the generated lyrics to become overly flashy and less natural.

\subsection{Musical Translation Test Dataset}
\label{appx:test_data}
\begin{figure}[t]
  \centering
  \includegraphics[width=1\linewidth]{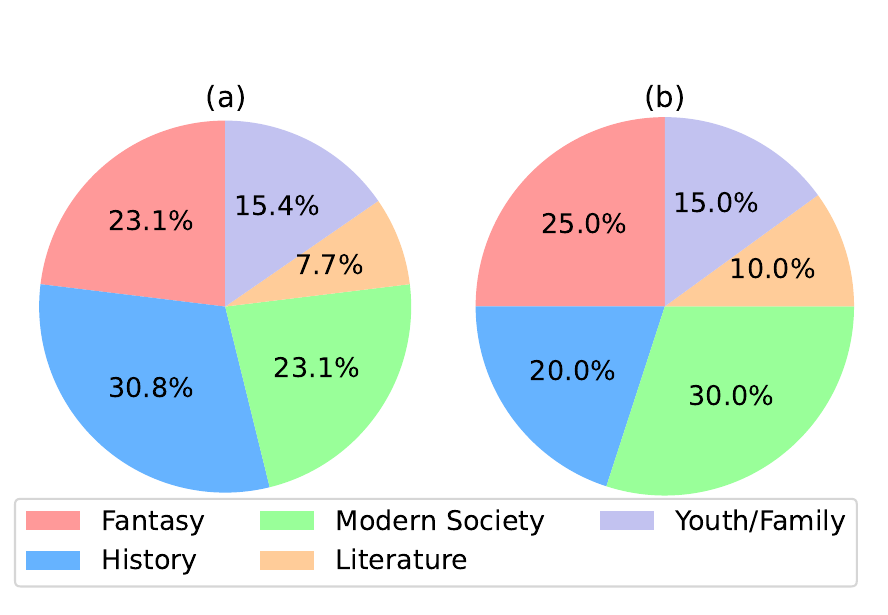}
  \vspace{-0.7cm}
  \caption{The distribution of musicals in \texttt{MusicalTransEval} dataset (a) and musical testing dataset (b).
  }
  \label{fig:annotation_distribution}
\vspace{-0.5cm}
\end{figure}

We manually collect the lyrics from \href{https://music.163.com/}{Cloud Music} and split them into paragraphs. The length constraint is obtained by counting the syllables of the English lyrics using the \href{https://github.com/mholtzscher/syllapy}{Syllapy library}. 
For testing BLEU and COMET scores, we collect the gold reference from human translations provided in Cloud Music.
Our final musical dataset consists of 409 paragraphs and 1742 lines and mainly serves the purpose of testing performance. The musical distribution is shown in Figure~\ref{fig:annotation_distribution}
(b). This test dataset is used to evaluate sentence-level and whole-song translation in the paper.

\section{Implementation details}
    \label{appendixB}
    \begin{table*}[htbp]
\fontsize{9.5}{9.5}\selectfont
  \centering
    \begin{tabular}{lp{30em}}
    \toprule
    \textbf{Model} &\textbf{Prompt}\\
    \midrule
    \textbf{Basic Reward Model} & 
You are a translation grader. Given English lyrics and a corresponding Chinese translation, you need to give scores in the range of 1-4 (4 is the highest) considering both fluency and translation accuracy.  Here are the metrics:\newline{}
Score 1: Not very fluent. There are inappropriate or awkward phrases or other big flaws.\newline{}
Score 2: Quite fluent, but there are serious translation mistakes that need correction.\newline{}
Score 3: Quite fluent, no big mistake in translation. But there are still small mistakes in phrasing or the translation of idioms.\newline{}
Score 4: Very fluent, no mistakes, and excellent translation.\newline{}
Note that a score of 4 means excellent and should be only given if you are absolutely sure the translated sentence is perfect. Any tiny mistake will make its score less than 4.\newline{}
Now, I will provide you with the English lyrics and the Chinese translation. You need to give me only one number and nothing else. For a comprehensive understanding, I will provide you the context: [paragraph].\newline{}
The English lyrics is: [original lyrics].\newline{}
The Chinese translation is: [translation]. The score is: \\
    \midrule
    \textbf{Advanced Reward Model} & 
You are a translation grader. Given a Chinese translation of lyrics, you need to give scores in the range 1-4 (4 is the highest) for whether it looks like good lyrics. Criteria for scoring:\newline{}
Score 1: The translation does not resonate as good lyrics.\newline{}
Score 2: Acceptable as lyrics, but mundane and unremarkable.\newline{}
Score 3: Good fit for lyrics with some literary flair and aesthetic language.\newline{}
Score 4: Outstanding lyrical quality, inventive, expressive, and captivating.\newline{}
Reserve a score of 4 for truly impressive lyricism and be prudent when giving 4. Regular conversational phrases typically merit a score of 2.\newline{}
Now, I will provide you with the Chinese translation. You need to give me only one number and nothing else. The Chinese translation is: [translation].\newline{}
The score is: \\
    \midrule
    \textbf{Translation Model w/o Rhyme} & 
I will give you an English lyric and you need to translate it into Chinese with exactly [length] characters. Please only output the translated results and nothing more. The English lyrics are: [original lyrics]. Then the translation result is: \\
    \midrule
    \textbf{Translation Model w/ Rhyme} & 
I will give you an English lyric and you need to translate it into Chinese with exactly [length] characters, where the ending rhyme type is [rhyme]. Please only output the translated results and nothing more. The English lyrics are: [original lyrics]. Then the translation result is: \\
    \bottomrule
    \end{tabular}%
  \caption{Prompts used for our two reward models and the translation model. For the translation model, we either only incorporate the length constraint or additionally add the rhyme constraint.}
  \label{tab:appx-prompt}%
\end{table*}%

\begin{table*}[htbp]
\fontsize{9.5}{9.5}\selectfont
  \centering
    \begin{tabular}{lp{30em}}
    \toprule
    \textbf{Model} &\textbf{Prompt}\\
    \midrule
    \textbf{GPT-4o paragraph translation} & 
Please translate the following English paragraph into Chinese, adhering strictly to the specified number of Chinese characters per line (commas do not count towards the character count). Maintain a strict line-by-line correspondence between the English and Chinese versions, ensuring the number of lines remains the same. Some examples: \newline{}
Example 1: \newline{}
The English paragraph is: You are sixteen going on seventeen,  \newline{}
baby it's time to think \newline{}
Better beware, be canny and careful,  \newline{}
baby you're on the brink \newline{}
The required character count for each line is: [10, 8, 9, 9] \newline{}
The translated version is: \newline{}
\begin{CJK}{UTF8}{gbsn}你十六岁，即将要十七岁\end{CJK} \newline{}
\begin{CJK}{UTF8}{gbsn}宝贝呀，该去思考了\end{CJK} \newline{}
\begin{CJK}{UTF8}{gbsn}应当警觉、谨慎和当心，\end{CJK} \newline{}
\begin{CJK}{UTF8}{gbsn}宝贝，你就在危险边缘\end{CJK} \newline{}
Example 2: \newline{}
The English paragraph is: I am I, Don Quixote, the Lord of La Mancha \newline{}
My destiny calls and I go \newline{}
And the wild winds of fortune will carry me onward \newline{}
Oh, whithersoever they blow \newline{}
Whithersoever they blow, onward to glory I go \newline{}
The required character count for each line is: [13, 7, 10, 6, 11] \newline{}
The translated version is: \newline{}
\begin{CJK}{UTF8}{gbsn}正是我，堂吉诃德，拉曼查的英豪\end{CJK} \newline{}
\begin{CJK}{UTF8}{gbsn}我的宿命召我前进\end{CJK} \newline{}
\begin{CJK}{UTF8}{gbsn}幸运的狂飙会策我向前\end{CJK} \newline{}
\begin{CJK}{UTF8}{gbsn}任他风吹雨打\end{CJK} \newline{}
\begin{CJK}{UTF8}{gbsn}任他风吹雨打，向荣誉进发！\end{CJK} \newline{}
Example 3: \newline{}
The English paragraph is: Even when the dark comes crashing through \newline{}
When you need a friend to carry you \newline{}
And when you're broken on the ground \newline{}
You will be found \newline{}
The required character count for each line is: [12, 10, 7, 5] \newline{}
The translated version is: \newline{}
\begin{CJK}{UTF8}{gbsn}即使当你的世界被黑暗吞没\end{CJK} \newline{}
\begin{CJK}{UTF8}{gbsn}当你需要朋友携手同行\end{CJK} \newline{}
\begin{CJK}{UTF8}{gbsn}当你摔落在地面\end{CJK} \newline{}
\begin{CJK}{UTF8}{gbsn}你会被发现\end{CJK} \newline{}
Example 4: \newline{}
The English paragraph is: Just because you find that life's not fair, \newline{}
it doesn't mean that you just have to grin and bear it! \newline{}
If you always take it on the chin and wear it \newline{}
Nothing will change. \newline{}
The required character count for each line is: [11, 15, 9, 7] \newline{}
The translated version is: \newline{}
\begin{CJK}{UTF8}{gbsn}若你只是觉得生活不公平，\end{CJK} \newline{}
\begin{CJK}{UTF8}{gbsn}那并不意味着你必须要微笑着忍受\end{CJK} \newline{}
\begin{CJK}{UTF8}{gbsn}如果你总是忍气吞声\end{CJK} \newline{}
\begin{CJK}{UTF8}{gbsn}没有事情会改变\end{CJK} \newline{}
Example 5: \newline{}
The English paragraph is: and do I dream again? \newline{}
for now I find \newline{}
the phantom of the opera is there \newline{}
inside my mind \newline{}
The required character count for each line is: [7, 5, 8, 4] \newline{}
The translated version is: \newline{}
\begin{CJK}{UTF8}{gbsn}我是否又做梦了\end{CJK} \newline{}
\begin{CJK}{UTF8}{gbsn}因为我发现\end{CJK} \newline{}
\begin{CJK}{UTF8}{gbsn}歌剧魅影就在那里\end{CJK} \newline{}
\begin{CJK}{UTF8}{gbsn}在我心中\end{CJK} \newline{}
The English paragraph is: [paragraph] \newline{}
The required character count for each line is: [the list of length constraints] \newline{}
IMPORTANT: Output ONLY the translated Chinese paragraph. Do not include any explanations, notes, or additional text. Your translation must strictly follow the given format and character counts. \newline{}
The translated version is: \\
    \bottomrule
    \end{tabular}%
  \caption{The prompt used for a simple method that directly prompts GPT-4o to translate an English lyrics paragraph, without or with the rhyme constraint. The 0-shot version is derived by directly deleting the five examples and thus is not displayed for simplicity.}
  \label{tab:gpt-paragraph-prompt}%
\end{table*}%

\noindent\textbf{Reward Model Training Details.} We use \texttt{Chinese-Alpaca-2-13B} for training reward models. See Table \ref{tab:appx-prompt} for detailed prompts used for our two reward models.

For our basic translation quality reward model, there are 471, 322, 971, and 2174 data samples with scores from 1 to 4. We upsample class 2 with a ratio of 1.5, downsample class 3 with a probability of 0.7, and downsample class 4 with a probability of 0.5. After adjusting the training dataset, we train our model with 5 epochs. Data downsampling means we keep each data sample with some probability, and data upsampling with a ratio $p$ means we first keep one copy of the dataset and then conduct data downsampling with probability $p-1$ to derive additional data samples.

For our advanced translation quality reward model, there are 3104 samples with label 2 and 834 samples with label 3. We downsample class 2 with a probability of 0.4, upsample class 3 with a ratio of 1.5, and then train 5 epochs.

\noindent\textbf{Translation Model Training Details.} We also use \texttt{Chinese-Alpaca-2-13B} as the translation base model. See Table \ref{tab:appx-prompt} for the prompts used for training. Both the two versions have the length constraint but one of them additionally has the rhyme constraint and is used in the second stage of the inference-time optimization framework. During translation model training, we mix the two prompts in the dataset so each data item appears twice (one with and the other without the rhyme constraint in the prompt).

We use 1 epoch for both training stages. Training on 1.75M data samples takes about 9 hours using 8 80GB A100 GPUs. The codebase is adopted from the \href{https://github.com/eric-mitchell/direct-preference-optimization}{DPO GitHub repository}~\cite{rafailov2023direct}, which also supports supervised fine-tuning. We use the training batch size of 32 and keep all other hyper-parameters default choices in that repository.

\noindent\textbf{Inference-time loss function Details.} We explain details in the inference-time loss function here: \begin{align*}
    &\mathcal L(y_1,\dots,y_n)=\sum_i\left(\lambda_1 [\operatorname{Rhy}(y_i)\ne \operatorname{Rhy}(y_n)]\right.\\
    &\left.+\lambda_2 D(\mathrm{gt}_i,|y_i|)-\lambda_3 R_{\mathrm{adv}}(y_i)-\lambda_4 R_{\mathrm{bas}}(y_i)\right).
\end{align*}

The penalty coefficient in function $D(\cdot,\cdot)$ is set as $\beta=2$. and the four hyperparameters are $$\lambda_1=2, \lambda_2=3, \lambda_3=1, \lambda_4=1.$$ According to our rubrics, the translation basic quality is a compulsory requirement to ensure acceptable translation results, we thus only consider those with $R_{\mathrm{bas}}\ge 3$ to ensure translations are preferable. We choose a temperature $T=0.7$ for generation and $\text{top-}p=0.95$. We may change to other hyperparameters to gain slightly better results, but in practice, this configuration can already achieve decent translation results.

Our pipeline with 40 + 40 samples runs within 8 hours on our musical test set and roughly takes 1 minute for each paragraph. In terms of real-world musical lyrics translation application, this speed is acceptable, thus during experiments we mainly focus on performance.

\section{Additional qualitative results}
    \label{appendixC}
    \begin{table*}[]
\fontsize{9}{10}\selectfont
\centering
\resizebox{\textwidth}{!}{%
\begin{tabular}{ccc}
\toprule
\textbf{Original lyrics} & 
\textbf{Ours \textsc{Ver.3} with reward model} &
\textbf{Ours \textsc{Ver.3} without reward model} \\
\midrule
Still strove, with his last ounce of courage, & 
\begin{CJK}{UTF8}{gbsn}还在\uwave{竭尽全力}地奋斗\end{CJK} & 
\begin{CJK}{UTF8}{gbsn}拼了命的继续着\underline{奋搏}\end{CJK} \\
To reach the unreachable stars! & 
\begin{CJK}{UTF8}{gbsn}要飞到最远的星宿\end{CJK} & 
\begin{CJK}{UTF8}{gbsn}去到那\underline{不曾到过的}\end{CJK} \\

\hline

Well, let that lonely feeling wash away &
\begin{CJK}{UTF8}{gbsn}让那寂寞的感觉冲刷开\end{CJK} &
\begin{CJK}{UTF8}{gbsn}让那孤独感觉洗刷\underline{一洗}\end{CJK} \\

Maybe there's a reason to believe you'll be okay &
\begin{CJK}{UTF8}{gbsn}也许有一些理由相信你会好起来\end{CJK} &
\begin{CJK}{UTF8}{gbsn}或许有理由相信你会过得很顺利\end{CJK}\\

Cause when you don't feel strong enough to stand &
\begin{CJK}{UTF8}{gbsn}当你感觉无力站起身来\end{CJK} &
\begin{CJK}{UTF8}{gbsn}\underline{因为}你太懦弱无法站立\end{CJK} \\
You can reach, reach out your hand&
\begin{CJK}{UTF8}{gbsn}你可以伸出手来\end{CJK} &
\begin{CJK}{UTF8}{gbsn}伸出你的手\underline{可以}\end{CJK}\\
\bottomrule
\end{tabular}
}
\caption{The effects of using reward model terms in optimization pipeline. Translational errors and awkward phrases are \underline{underlined}. Excellent lyrics are \uwave{underwaved}}
\vspace{-0.6cm}
\label{tab:qualitative-reward}
\end{table*}

Table~\ref{tab:qualitative-reward} showcases the qualitative effect of using reward models in the optimization framework. Without reward model terms, the translation quality significantly drops. Additional translation results are shown in Table~\ref{tab:more-demo}. 

\begin{CJK}{UTF8}{gbsn}

\begin{table*}[]
\centering
\begin{tabular}{cc}
\toprule
\textbf{Original lyrics} & 
\textbf{Translation results}\\
\midrule
I am I, Don Quixote, the Lord of La Mancha & 
我是我，堂吉诃德，拉曼查的领主 \\
My destiny calls and I go &
宿命呼唤，我随之去\\
And the wild winds of fortune, will carry me onward&
命运的狂风，将我带向\uwave{未知旅途}\\
Oh whithersoever they blow &
任凭风儿吹向何处\\
Whithersoever they blow, onward to glory I go&
任凭风向何处吹，我直奔荣耀而去\\

\hline

Hamilton faces an endless uphill climb & 
汉密尔顿面临无尽上坡路 \\
He has something to prove & 
他要证明什么\\
He has nothing to lose & 
他\uwave{无后顾之忧}\\
Hamilton’s pace is relentless & 
汉密尔顿\uwave{步履不停}\\
He wastes no time &
他\uwave{毫不耽搁}\\
What is it like in his shoes? &
\uwave{他脚下的路如何}？\\

\hline
So let the sun come streaming in&
就让阳光\uwave{洒满房间}\\
Cause you'll reach up and you'll rise again&
因为你会奋起\uwave{再登攀}\\
Lift your head and look around&
抬起头四处看看\\
You will be found&
必被发现\\

\hline
you will be popular!&
你会受到欢迎\\
You're gonna be popular! & 
你将会很有人气\\
I'll teach you the proper poise&
姿势得体我来教\\
When you talk to boys&
与男生谈笑\\
Little ways to flirt and flounce&
小动作挑逗撒娇\\

\hline
To dream the impossible dream,&
追求不可能的梦想 \\ 
To fight the unbeatable foe,&
挑战\uwave{不可战胜之敌} \\ 
To bear with unbearable sorrow,&
承受那难以承受之痛 \\ 
To run where the brave dare not go&
\uwave{勇闯}无人敢去之地\\

\hline
I wrote my way out&
我\uwave{以笔自救} \\
Wrote everything down far as I could see&
写下所见所闻，\uwave{尽我所能} \\
I wrote my way out&
我写下出路 \\
I looked up and the town had its eyes on me&
我抬头，全镇都在注视着我 \\

\bottomrule
\end{tabular}

\caption{More qualitative results of our method. Excellent lyrics are \uwave{underwaved}}
\label{tab:more-demo}
\end{table*}

\end{CJK}


\begin{figure*}[ht]
  \centering
  \includegraphics[width=1\linewidth]{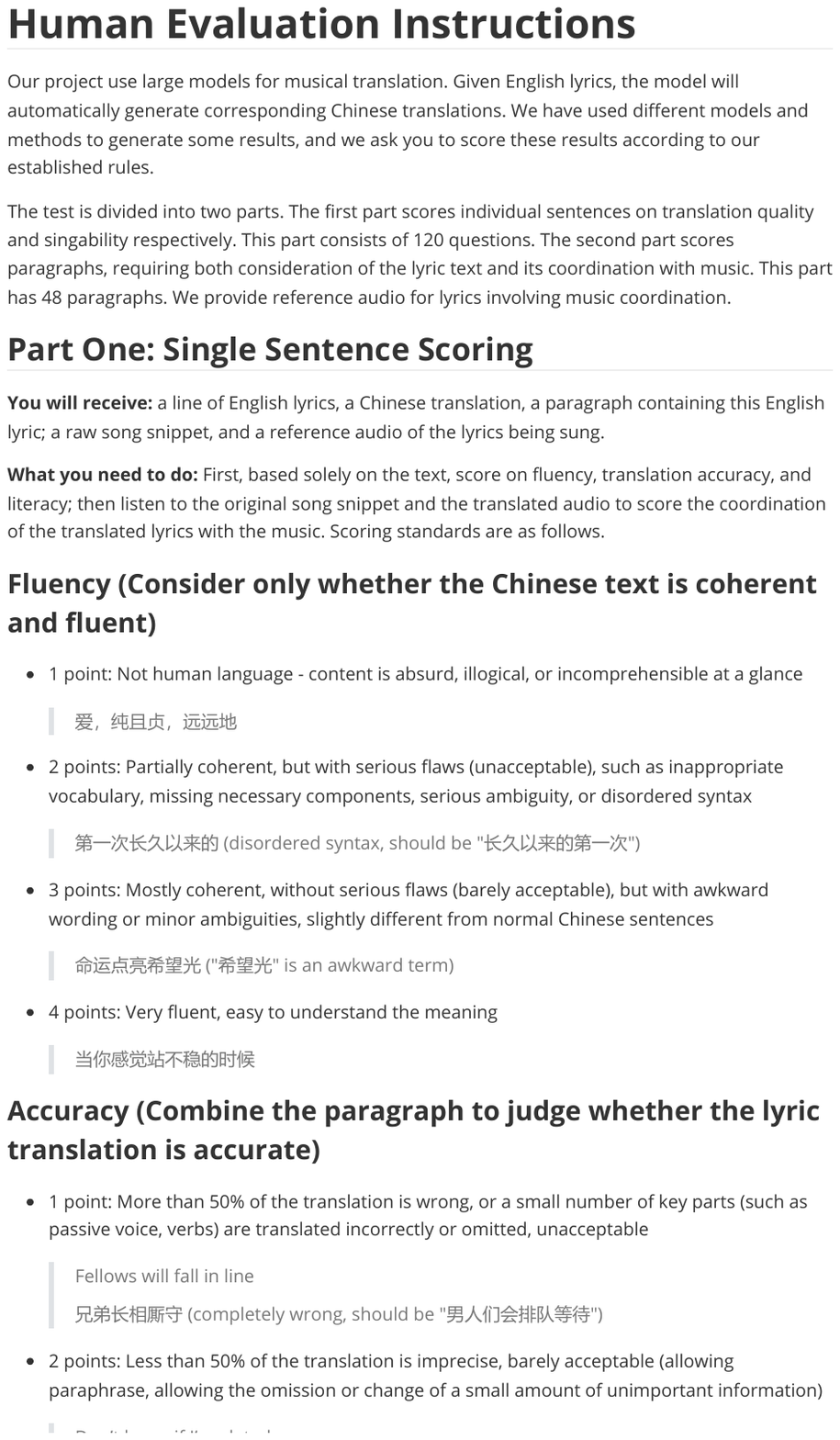}
  \vspace{-0.1cm}
  \caption{Instructions for human evaluation, page 1/3.}
  \label{fig:instruction1}
\vspace{-0.1cm}
\end{figure*}
\begin{figure*}[ht]
  \centering
  \includegraphics[width=1\linewidth]{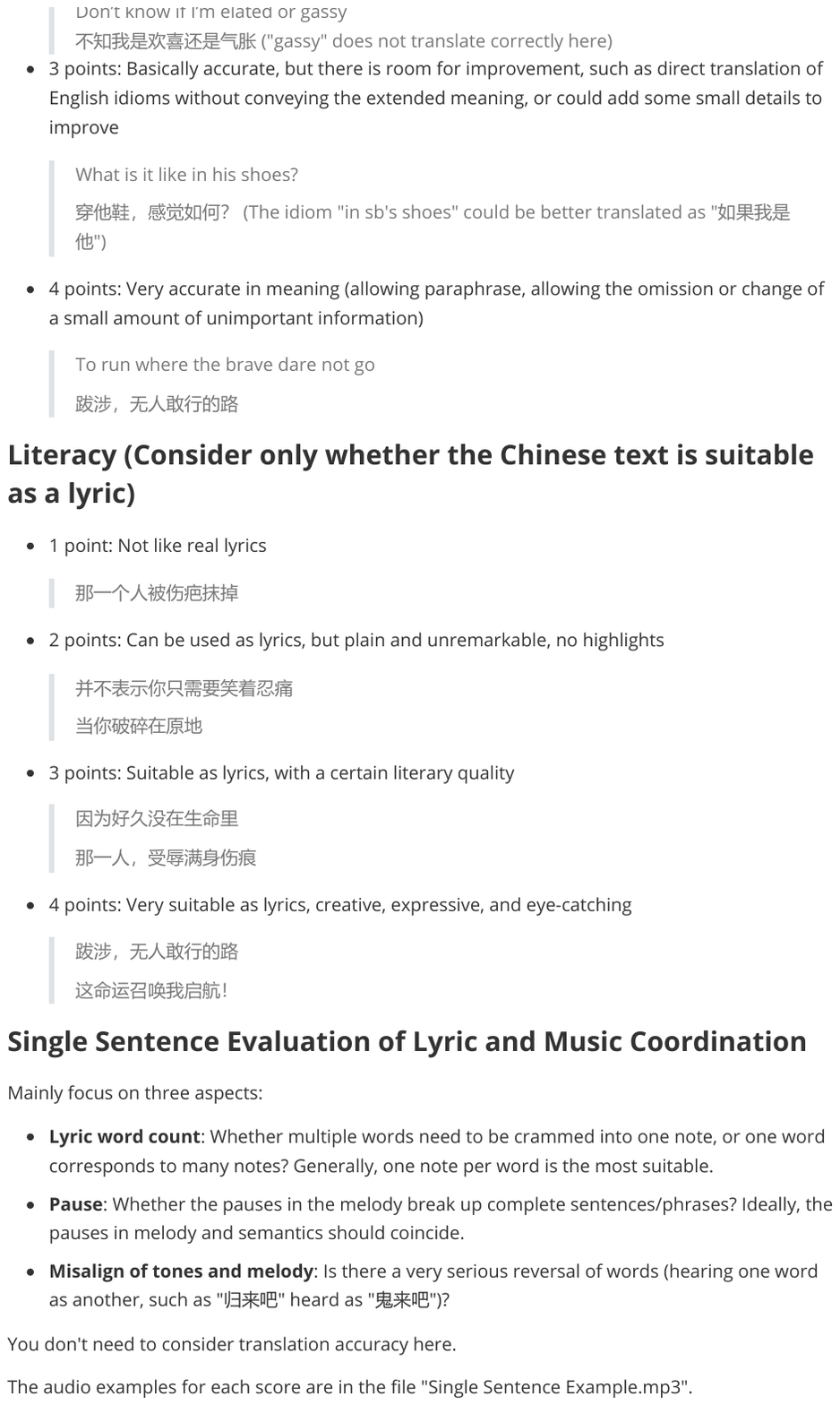}
  \vspace{-0.1cm}
  \caption{Instructions for human evaluation, page 2/3.}
  \label{fig:instruction2}
\vspace{-0.1cm}
\end{figure*}
\begin{figure*}[ht]
  \centering
  \includegraphics[width=1\linewidth]{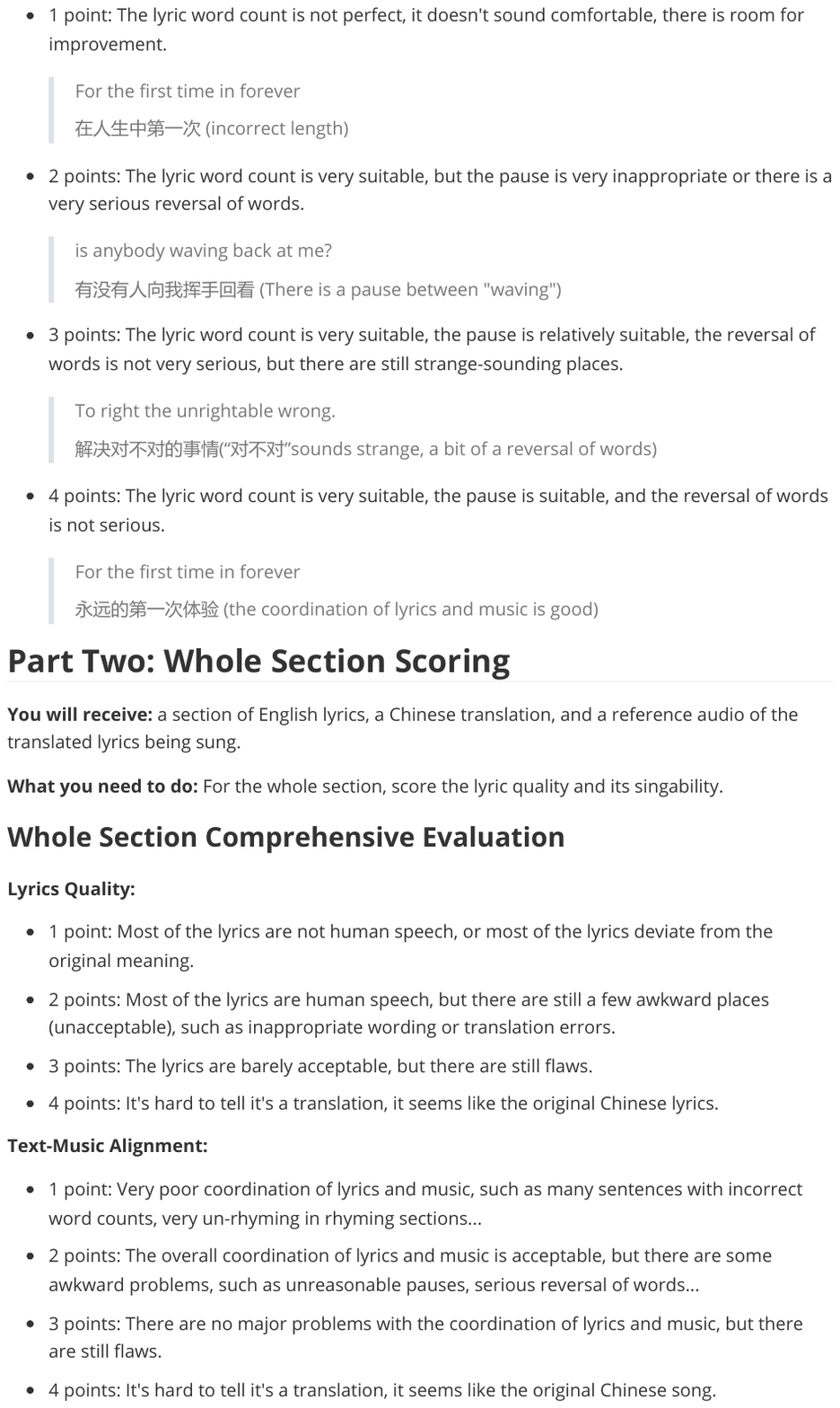}
  \vspace{-0.1cm}
  \caption{Instructions for human evaluation, page 3/3.}
  \label{fig:instruction3}
\vspace{-0.1cm}
\end{figure*}

\section{Human Evaluation Details}
    \label{appendixD}

We recruited 4 local college students who are musical enthusiasts from the college's musical club. We randomly sampled 30 sentences and 12 paragraphs from our test set, allowing the baseline and three versions of our model to generate 120 sentences and 48 paragraphs. We then asked another musical enthusiast to sing all the generated results.
The evaluators assigned scores for fluency, accuracy, literacy, and music-text alignment for the sentence results, and overall translation quality and music-text alignment for the paragraph results. We provided detailed scoring rubrics with examples and required the participants to adhere to our rules. The English version of the instructions is shown in Figures~\ref{fig:instruction1}, \ref{fig:instruction2}, and \ref{fig:instruction3}.
Each annotator took 3 hours to complete the evaluations, and we compensated them with a reasonable price for university students.

To test the reliability of our human evaluations,
we computed inter-rater agreement using intraclass coefficients (two-way mixed-effect, average measure model), following the practice of~\citet{ou-etal-2023-songs}. The results are as follows: 0.681 for sentence-level fluency, 0.727 for sentence-level accuracy, 0.546 for sentence-level literacy, 0.485 for sentence-level music-text alignment, 0.664 for paragraph-level overall translation quality, and 0.498 for paragraph-level music-text alignment. According to~\citet{koo2016guideline}, most of them fall into the ``moderate reliability'' range (0.5 to 0.75).

\end{document}